\PassOptionsToPackage{numbers,sort&compress}{natbib} \documentclass{article}
\usepackage[preprint]{corl_2026} 
\usepackage{graphicx}
\usepackage{amsmath}
\usepackage{amssymb}
\usepackage{booktabs}
\usepackage{float}
\usepackage{bm}
\usepackage{diagbox}
\usepackage{makecell}
\usepackage{booktabs}

\title{KEMO: \underline{E}vent-Driven \underline{K}eyframe \underline{M}em\underline{o}ry for Long-Horizon
Robot Manipulation with VLA Policies}

\author{
\textbf{Yihan Zeng}$^{1,3,4}$, \quad
\textbf{Minghao Ye}$^{3,5}$, \quad
\textbf{Yiyuan Chen}$^{3}$, \quad \\
\textbf{Yide Shentu}$^{3}$\textbf{,} \quad
\textbf{Philipp Wu}$^{3}$\textbf{,} \quad
\textbf{Zike Yan}$^{1,2}$\textbf{,} \quad
\textbf{Zhongyu Li}$^{1,2}$ \\[0.6em]
\normalfont
$^{1}$Hong Kong Embodied AI Lab
\quad
$^{2}$The Chinese University of Hong Kong
\quad
$^{3}$xdof.ai \\
\quad
$^{4}$University of Electronic Science and Technology of China
\quad
$^{5}$Shanghai Jiao Tong University\\[0.4em] 
Emails: \texttt{zengyh@std.uestc.edu.cn,zhongyuli@cuhk.edu.hk}, Website: https://Hatty-z.github.io/KEMO
}

\begin{document}
\maketitle

\begin{figure}[H]
\centering
\includegraphics[width=1\linewidth]{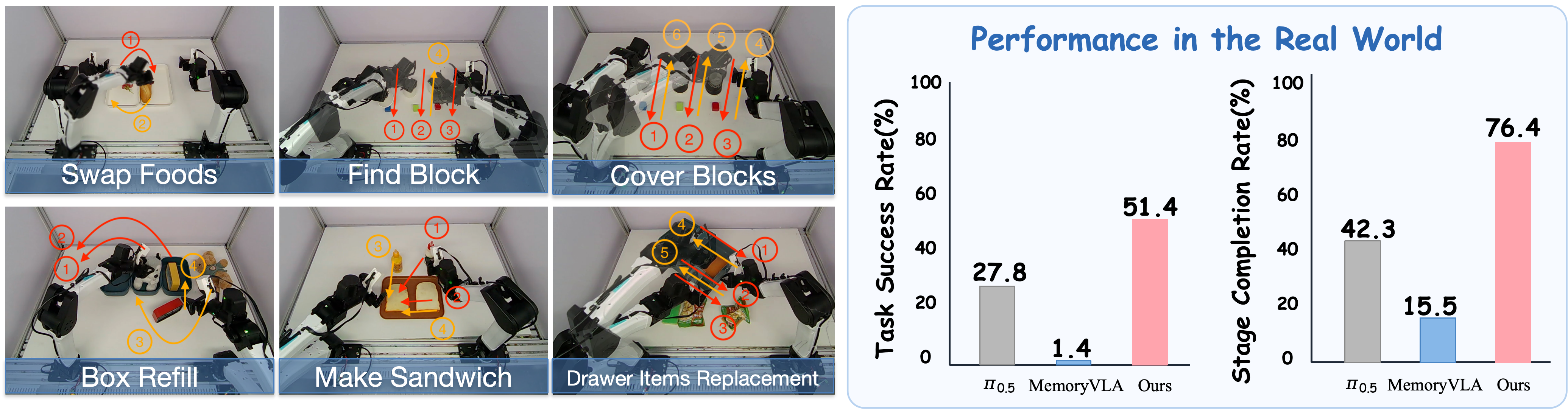}
\caption{We propose a plug-in event-driven keyframe memory framework that detects task-relevant event keyframes from robot states and visual observations, and stores the selected keyframes in a memory bank to provide history evidence for VLA policies. Left: Various memory-dependent dual-arm manipulation tasks realized by our method in the real world, spanning average trajectory durations of 28 to 95 seconds and average trajectory lengths of 830 to 2846 steps. Right: Aggregate Task Success Rate (TSR) and Stage
Completion Rate (SCR) across the six tasks (12 trials per task), comparing KEMO with
memory-free and memory-augmented baselines.}
\label{fig:introduce}
\end{figure} 
\begin{abstract}
      Long-horizon robot manipulation remains challenging because similar observations may occur at different execution stages, while the appropriate action depends on previously completed operations. Memory can address this ambiguity by enabling policies to infer task progress from execution history. However, existing memory-augmented approaches often either retain dense histories that require compression or rely primarily on recent context that may discard earlier task-relevant events. In this work, we propose propose \textbf{KEMO}, a lightweight plug-in memory framework that automatically selectively preserves keyframes associated with task-relevant state changes for VLA policies. KEMO combines robot kinematics with visual filtering to detect events, encodes the selected keyframes as compact temporally ordered memory tokens, and integrates them with current visual features through cross-attention and gated residual fusion for VLA training. The detected events also define higher-weight training samples near critical transitions. We evaluate KEMO on various real-world dual-arm manipulation tasks spanning 2 to 6 scored subtasks, and trajectory length ranging from 830 steps to 2846 execution steps (durations from 28 to 95 seconds). Compared with the memory-free baseline (\textit{e.g.}, $\pi_{0.5}$), KEMO improves aggregate Task Success Rate by 23.6\% and Stage Completion Rate by 34.1\%. Ablations show that event-driven keyframe selection outperforms uniform sampling and recent-frame retention, while the proposed gated fusion and keyframe-aligned loss weighting provide complementary gains.

\end{abstract}

\keywords{Robot Learning, Long Horizon Task And Motion Planning} 


\section{Introduction}
	
Long-horizon robot manipulation requires executing sequences of interdependent subtasks over extended time horizons. In such tasks, robots often face stage ambiguity and partial observability: visually similar observations may appear at different execution stages, while the correct action depends on what has already happened. Vision-Language-Action (VLA) policies ~\citep{intelligence2025pi05visionlanguageactionmodelopenworld,kim2024openvlaopensourcevisionlanguageactionmodel,brohan2023rt2visionlanguageactionmodelstransfer,chi2024diffusionpolicyvisuomotorpolicy,kim2025finetuningvisionlanguageactionmodelsoptimizing} that condition only on the current observation therefore struggle to determine the proper action at these disambiguation points.

A natural solution is to equip policies with memory of past execution history to provide evidence of completed task stages and previously observed
information, thereby disambiguating visually similar current states. Existing memory-augmented VLA methods mainly follow two directions. Dense memory methods store all the past or recent frames, but must compress history through token merging \cite{shi2025memoryvla}, recurrent states \cite{dai2026robommebenchmarkingunderstandingmemory}, or fixed-size latent representations \cite{li2025cronusvlaefficientrobustmanipulation}, which may mix sparse task-state transitions with large amounts of redundant observations and make critical events less distinguishable in the compressed memory representation. Task-decomposition methods use Vision-Language Models (VLMs) or high-level modules \cite{sridhar2025memer} \cite{chen2026rmbenchmemorydependentroboticmanipulation} to track progress. However, these methods may require subtask-level supervision, introduce additional online inference cost, and propagate errors from generated semantic summaries or subgoals from the task decompositors. These limitations motivate a lightweight memory mechanism that directly preserves sparse task-relevant events, without compressing dense observation histories or relying on explicit task decomposition and additional online high-level reasoning.

We propose \textbf{KEMO}, a plug-in memory framework for long-horizon manipulation. Our key insight is that event keyframes, corresponding to task-relevant state transitions, provide compact and explicit evidence of execution history, which could disambiguate visually similar observations and support stage-appropriate action prediction. We detect such keyframes by combining kinematic cues with visual change verification: motion-based saliency identifies candidate transition moments, while a visual deduplication filter removes candidates that do not correspond to meaningful scene changes. The selected keyframes are encoded into a compact memory bank and fused with the current visual representation through cross-attention, enabling the VLA policy to predict actions using both current observations and historical transition evidence. The same keyframes are also used to define high-weight supervision windows during training, aligning memory construction with learning emphasis at critical moments.

We evaluate KEMO on various challenging long-horizon dual-arm manipulation tasks in the real world, spanning from 2 to 6 scored subtasks and average trajectory durations from 28 to 95 seconds. Across all tasks, KEMO improves aggregate Task Success Rate (TSR) by 23.6\% and average Stage Completion Rate (SCR) by 34.1\% over the memory-free $\pi_{0.5}$ baseline. Compared with MemoryVLA, KEMO improves TSR by 50.0\% points and SCR by 60.9\%.  Event-driven keyframe selection also outperforms alternative memory selection strategies such as uniform sampling and recent-frame retention. On the Cover Blocks task, removing gated memory fusion decreases TSR and SCR by 75.0\% and 87.5\%, respectively, while removing keyframe-aligned loss weighting decreases them by 50.0\% and 45.8\%. These results isolate the complementary contributions of event-based memory selection, gated fusion, and transition-aligned supervision.

In summary, our contributions are three-fold: (1) \textbf{A novel plug-in event-driven keyframe memory framework for long-horizon manipulation using VLA policies.}
We introduce a keyframe memory module that stores task-relevant keyframes and injects compressed keyframe tokens into the visual representation, providing explicit history evidence to existing VLA policies. (2) \textbf{A new task-agnostic keyframe detector based on robot states and visual observations.} We propose a lightweight detector that identifies stage-critical event moments by combining robot motion patterns with visual change verification, without requiring task-specific stage annotations or additional online
VLM task decomposition which may introduce latency and propagated hallucinations. (3) \textbf{Strong empirical results on real-world long-horizon tasks.}  We evaluate our method on six challenging long-horizon real-robot tasks with 2 to 6 subtasks, trajectory durations ranging from 28s to 95s (trajectory length ranging from 830 steps to 2846 steps), demonstrating consistent improvements over memory-free and memory-augmented VLA baselines, with ablations validating the benefits of event-driven selection and keyframe-aligned supervision.


\section{Related Work}
Long-horizon manipulation with keyframe-based memory involves two closely related problems: (1) detecting task-critical moments as ``keyframes" from continuous robot trajectories, and (2) incorporating the selected keyframes into policy learning and action prediction. We review prior work along these two directions.

\textbf{Keyframe Detection for Manipulation.}
Keyframes provide a sparse, behaviorally meaningful abstraction of a manipulation trajectory, isolating moments of task-relevant transition from redundant intermediate frames. Existing approaches fall into two categories.
\textbf{VLM-based detection} reasons over visual content to identify keyframes. BPP~\cite{mark2026bpp} queries an off-the-shelf VLM to flag task-relevant events and conditions a downstream diffusion policy on the detected frames. MemER~\cite{sridhar2025memer} fine-tunes a video-VLM that jointly nominates keyframes and emits language subgoals, folding detection into the policy itself. While these methods capture semantic events with the strong reasoning capacity from VLMs, they introduce per-step high VLM inference costs and large model dependency,  and the hallucination from VLMs could propagate to the robot policies.
\textbf{Kinematic heuristics} instead mark a frame whenever the gripper toggles or joint velocities drop near zero, capturing grasps, releases, and stationary poses. Such rules are adopted by C2F-ARM~\cite{james2022coarse}, PerAct~\cite{shridhar2023perceiver}, SAM2Act~\cite{fang2025sam2act}, and TGM-VLA~\cite{pu2026tgm} for supervision extraction, and generalized by FrameSkip~\cite{yu2026frameskip} into a continuous importance score driven by action variation and gripper transitions. These rules are cheap, deterministic, and effective at capturing manipulation-critical moments.
Our method adopts kinematic cues for the same reasons. However, since they fire only on actor motion and produce redundant keyframes during repetitive gripper activity, we further introduce a visual signal to filter out uninformative frames and retain only those reflecting genuine task-relevant stage changes.

\textbf{Memory Augmented VLA Policies}.
VLA policies provide high-capacity visual-language representations for action prediction, making them natural backbones for incorporating temporal memory in long-horizon tasks.Equipping VLA models with memory has emerged as a direct approach to non-Markovian tasks. Methods differ primarily in memory granularity and representation.
\textbf{Dense and compressed memory} methods retain all or a fixed window of recent observations. ContextVLA~\cite{jang2025contextvla} compresses past frames into a single context token. HAMLET~\cite{koo2025hamlet} maintains a fixed window of time-contrastive moment tokens. ReMem-VLA~\cite{li2026remem} uses dual-level EMA recurrent queries updated at frame and chunk boundaries. RoboFlamingo~\cite{li2023vision} and SeedPolicy~\cite{gui2026seedpolicy} propagate history via LSTM and gated attention. MEM~\cite{torne2026memmultiscaleembodiedmemory} combines a video encoder with a self-updating language summary, while MemoryVLA~\cite{shi2025memoryvla} consolidates per-step features through token merging. As the observation history grows, these methods must either discard
older observations or compress them into a bounded representation, which may remove sparse decision-relevant information.
\textbf{Sparse keyframe memory} methods retain only semantically significant frames. BPP~\cite{mark2026bpp} prompts an off-the-shelf VLM to flag keyframes. SAM2Act+~\cite{fang2025sam2act} augments a manipulation transformer with a SAM2-inspired memory bank. Keyframe-Chaining VLA~\cite{chen2026non} learns a Keyframe Selection Module with FiLM task modulation. MemER~\cite{sridhar2025memer} fine-tunes a VLM to jointly select keyframes and emit language subgoals. Mem-0~\cite{chen2026rmbenchmemorydependentroboticmanipulation} keeps one frame per completed subtask for high-level planning, with short buffers at execution. Our method likewise builds memory from keyframes, fusing it with the current observation through masked cross-attention and gated residual fusion.

\section{Method}
\begin{figure}
\centering
\includegraphics[width=1\linewidth]{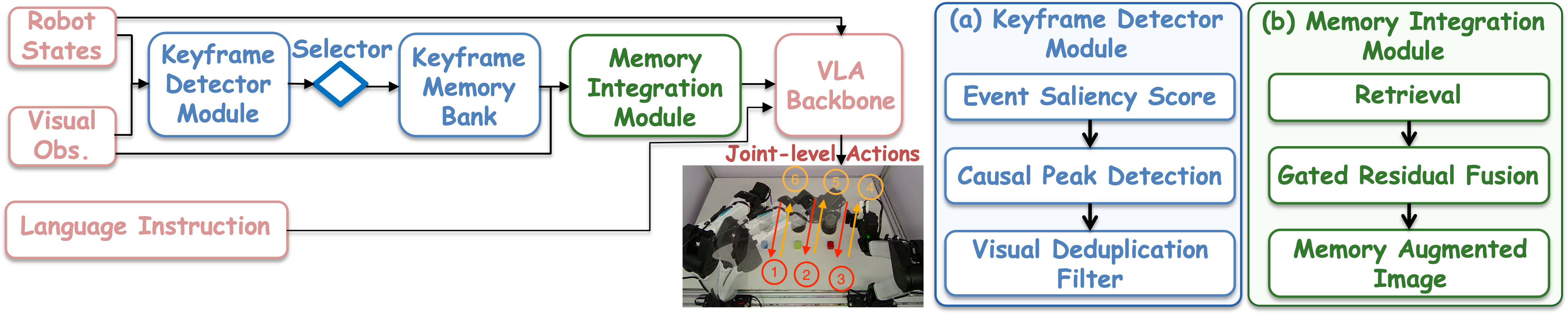}
\caption{\textbf{Event-driven keyframe memory framework.} The keyframe detector module uses robot states and visual observations to select task-relevant event keyframes, which are stored in a keyframe memory bank.
The memory integration module retrieves the stored keyframes and fuses them with the current visual representation, enabling the VLA backbone to predict actions using both current observations and historical transition evidence.
Right: module details for keyframe detection and memory integration.
}
\label{fig:overview}
\end{figure}
In this section, we introduce our KEMO framework, a plug-in framework that augments VLA policies with historical keyframe evidence for long-horizon robot manipulation. As shown in Fig.~\ref{fig:overview}, we posit that not all historical frames are equally useful: the most informative memories are event keyframes where robot actions produce task-relevant scene changes. Based on this principle, our framework detects event keyframes from kinematic and visual cues, stores them in a compact memory bank, and injects the memory into the current visual representation through cross-attention. We formulate the memory-conditioned policy in Sec.~\ref{sec:problem_formulation}, describe event-driven keyframe detection in Sec.~\ref{sec:keyframe_detection}, present keyframe memory integration in Sec.~\ref{sec:keyframe_memory}, and introduce keyframe-consistent training in Sec.~\ref{sec:keyframe_training}.
    
\subsection{Problem Formulation}
\label{sec:problem_formulation}

We consider long-horizon robot manipulation tasks characterized by stage ambiguity and partial observability. At each timestep $t$, the robot receives an observation $o_t = \{I_t, s_t\}$ consisting of color (RGB) images $I_t$ from multiple cameras and proprioceptive state $s_t \in \mathbb{R}^d$, together with a language instruction $\ell$, and then executes an action chunk $\mathbf{a}_{t:t+H} \in \mathbb{R}^{H \times D}$ over a prediction horizon $H$.

We define \emph{stage ambiguity} as the case where similar current observations correspond to different task states and therefore require different actions. This ambiguity arises because the current observation may not reveal which operations have already been completed or what task-relevant information was observed earlier. For example, as illustrated in Fig. \ref{fig:introduce},   similar observations in different stages of Cover Blocks require actions toward different target objects.

To resolve this ambiguity, the policy conditions on a history memory $\mathcal{M}_t$:
\begin{equation} \pi_{\theta} \left( \mathbf{a}_{t:t+H} \mid o_t,\ell,\mathcal{M}_t \right), \end{equation}
where $\mathbf{a}_{t:t+H}\in\mathbb{R}^{H\times D}$ denotes the $H$-step action chunk predicted from timestep $t$. Rather than retaining the complete observation history, we construct $\mathcal{M}_t$ as a temporally ordered set of event keyframes detected up to the current timestep $t$. Each keyframe corresponds to a task-relevant state transition induced by
the robot's interactions, such as covering, removing, placing, or replacing an object. The resulting memory provides compact historical evidence of completed task progress and previously observed task-relevant relationships for subsequent action prediction.

\subsection{Framework Overview}

As illustrated in Fig.~\ref{fig:overview}a, our method augments a VLA policy with an event-driven keyframe memory pipeline. At each timestep, the current multi-view visual observations and robot states are first processed by the keyframe detector. The detector uses kinematic cues to identify candidate transition moments and visual change to remove redundant candidates. Its
output is a sequence of accepted event keyframes together with their timestep indices.

The accepted keyframes are then encoded into compact visual tokens and stored in the temporally ordered memory bank $\mathcal{M}_t$. The memory integration module takes the current visual tokens as queries and the stored keyframe tokens as historical context. Through masked cross-attention and gated residual fusion, it produces a memory-augmented visual representation, which is subsequently combined with the current robot state and language instruction by the VLA backbone to predict the next action chunk.

The detected keyframe indices also provide a training signal. In addition to determining which observations are stored in $\mathcal{M}_t$, they define higher-weight supervision windows around task-relevant transitions. Consequently, keyframe detection provides both the memory content used for action prediction and the temporal signal used for keyframe-consistent training.

We describe the event-driven keyframe detector in Sec.~\ref{sec:keyframe_detection}. Its selected keyframes are used to construct and update the memory bank introduced in Sec.~\ref{sec:keyframe_memory}. The resulting memory tokens are fused with the current visual representation before action prediction, while the same keyframe indices are reused by the loss-weighting objective in Sec. \ref{sec:keyframe_training}.

\subsection{Event-Driven Keyframe Detection}
\label{sec:keyframe_detection}
\begin{figure}[t]
    \centering

    \begin{minipage}[t]{0.48\linewidth}
        \centering
        \includegraphics[width=\linewidth]{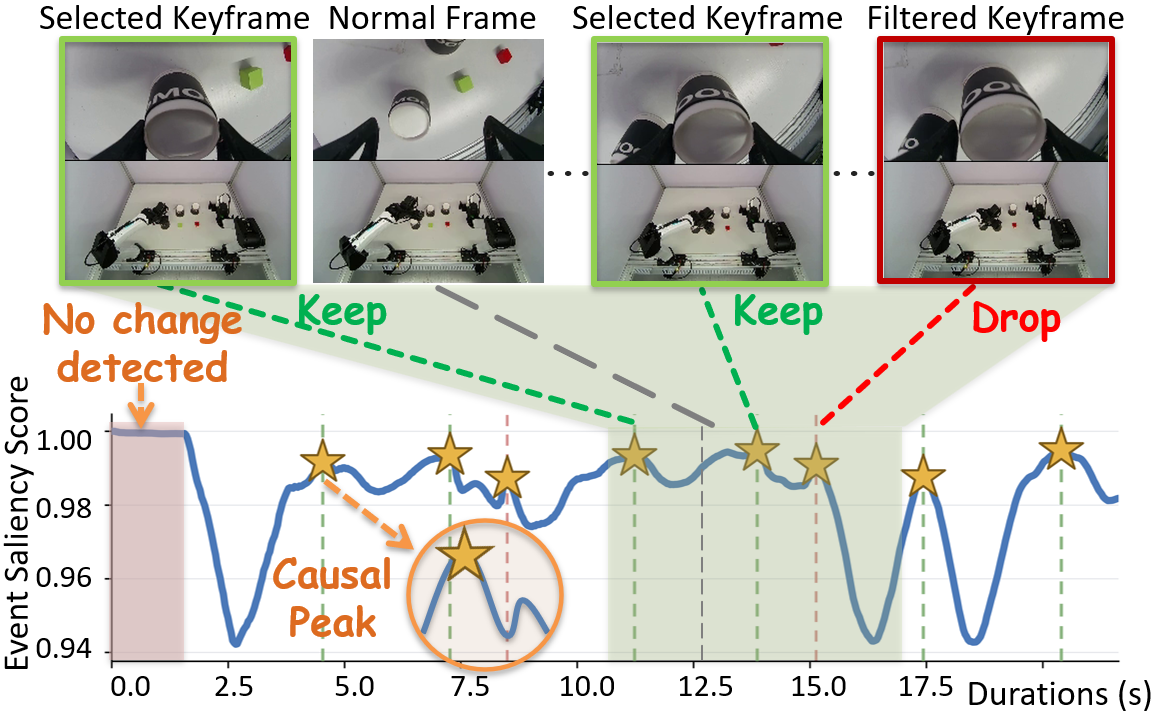}
        \small (a) Keyframe Detection 
    \end{minipage}
    \hfill
    \begin{minipage}[t]{0.48\linewidth}
        \centering
        \includegraphics[width=\linewidth]{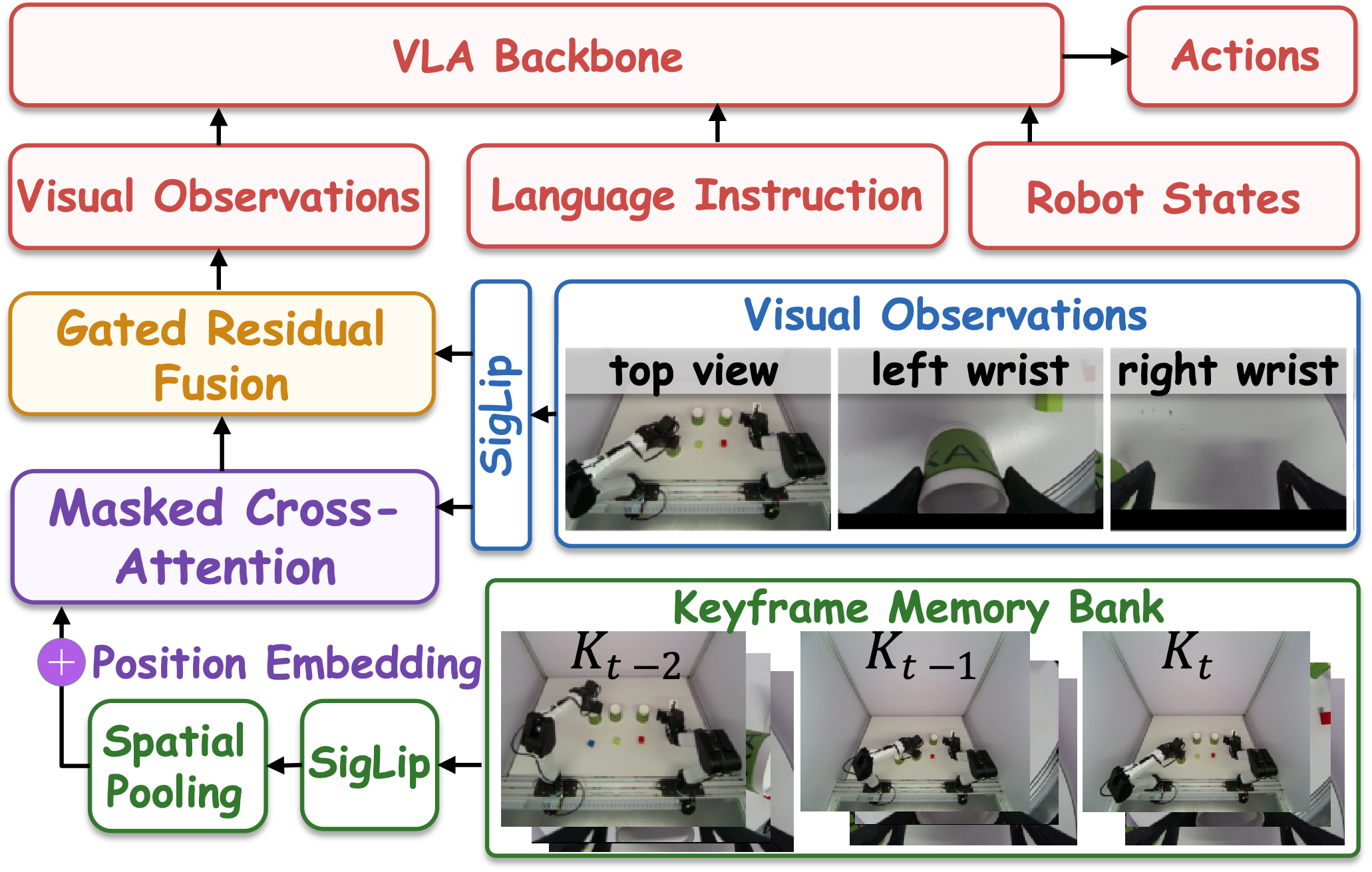}
        \small (b) Keyframe Memory Integration Module
    \end{minipage}

    \caption{
   \textbf{Event-driven keyframe detection and memory integration.}
(a) The keyframe detector computes an event saliency score from robot motion, identifies candidate keyframes through causal peak detection, and applies a visual deduplication filter to keep only candidates associated with meaningful scene changes. Selected keyframes are stored, while visually redundant or non-transition frames are filtered out.
(b) The memory integration module encodes stored keyframes with the visual encoder, applies spatial pooling and temporal positional embeddings, retrieves keyframe memory tokens, and fuses them with the current visual representation through masked cross-attention and gated residual fusion before action prediction by the VLA backbone.
    }
    \label{fig:keyframe_detection_memory}
\end{figure}
Reliably identifying stage transitions without task-specific annotation is the central challenge. As shown in Fig.~\ref{fig:keyframe_detection_memory}a, our method scores each timestep using kinematic signals and applies a visual filter to retain only semantically meaningful candidates.

\textbf{Event Saliency Score}. We define an \textit{event saliency score} based on recent motion features:
\begin{equation}
S_t = \frac{1}{1 + \bar{\delta}_t},
\end{equation}
where $\bar{\delta}_t = \frac{1}{w}\sum_{i=0}^{w-1}\|q_{t-i} - q_{t-i-1}\|_2$ is the mean joint-space displacement over a sliding window of length $w$, and $q_t$ denotes the robot joint-position vector at timestep $t$. The additive constant in the denominator prevents division by zero and yields the maximum score $S_t=1$ when no joint displacement is observed within the window. A high score indicates that the robot action has slowed down recently, which commonly occurs near key manipulation events such as grasping, placing, covering, or releasing an object, where deliberate deceleration ensures refined execution before transitioning to the next stage. 

\textbf{Causal Peak Detection}. Given the event saliency sequence ${S_t}$, we identify candidate keyframes as local peaks of $S_t$. During online execution, a candidate peak is confirmed after observing a short temporal window. These peaks correspond to moments where the robot motion locally slows down, yielding candidate event keyframes.

\textbf{Visual Deduplication Filter}. Motion-based keyframe candidates may also arise from pauses or adjustments that do not produce a meaningful change in the scene. We therefore compare each candidate frame $c_i$ with the most recently accepted keyframe $k_{i-1}$ using a frozen DINOv2 encoder~\cite{oquab2024dinov2learningrobustvisual}:
\begin{equation}
    v_i =
    1 -
    \frac{
        \phi(I_{c_i})^\top \phi(I_{k_{i-1}})
    }{
        \left\lVert\phi(I_{c_i})\right\rVert_2
        \left\lVert\phi(I_{k_{i-1}})\right\rVert_2
    },
\end{equation}
where $\phi(\cdot)$ denotes the global image embedding produced by the frozen encoder. The candidate is accepted as the next keyframe only if $v_i > \epsilon$, where $\epsilon$ is a predefined visual-dissimilarity threshold. Otherwise, it is discarded, and $k_{i-1}$ remains the reference for the next candidate. This step removes visually redundant motion candidates for which the observed scene remains largely unchanged.

Finally, we enforce a minimum refractory period of $r$ frames between consecutive accepted keyframes to avoid repeated detections within the same sustained transition.

\subsection{Keyframe Memory Integration}
\label{sec:keyframe_memory}
After detecting event keyframes, the next challenge is how to make them useful for action prediction. A keyframe only becomes effective memory if it can be represented compactly  to reduce redundant spatial information and computational overhead, ordered temporally, and fused with the current observation in a way that preserves the base VLA policy. We therefore design a memory integration module that encodes detected keyframes into compact visual tokens, stores them in a temporally ordered memory bank, and injects the memory into the current visual representation through cross-attention and gated fusion, shown in Fig.~\ref{fig:keyframe_detection_memory}b. 

\textbf{Keyframe Encoding}. Given the detected keyframe indices for each training trajectory, we encode keyframe images using the same SigLIP\cite{zhai2023sigmoidlosslanguageimage} visual encoder as the base policy. For each keyframe at index $k_i$, the encoder produces patch-level tokens:  $Z_{k_i} = \text{SigLIP}(I_{k_i}) \in \mathbb{R}^{N \times D}$  , where $N = 256$ is the number of spatial patches and $D$ is the embedding dimension. To reduce computational cost while preserving spatial structure, we apply $4{\times}4$ spatial average pooling to obtain $F_{k_i} \in \mathbb{R}^{16 \times D}$, reducing each keyframe from 256 to 16 tokens. 

\textbf{Keyframe Memory Bank Construction}. At timestep $t$, the memory bank $\mathcal{M}_t$ contains up to the $K$ most recent accepted keyframes whose indices satisfy $k_i \leq t$. When fewer than $K$ keyframes have occurred, the bank is padded with the most recent keyframe and a binary padding mask $\mathbf{m} \in \{0,1\}^K$ is maintained to suppress padded slots during attention. The slot-level mask is expanded over the 16 pooled tokens associated with each keyframe. We set $K$ according to subtask transitions of each task. 

\textbf{Temporal Positional Encoding}. To convey stage ordering, each keyframe slot receives a learnable temporal position embedding indexed by its position in the bank: $\tilde{F}_{k_i} = F_{k_i} + \text{PE}(i) \in \mathbb{R}^{16 \times D}$, where $\mathrm{PE}(i)$ is the learnable embedding of the $i$-th temporal slot, and is broadcast to all 16 pooled tokens of the corresponding keyframe. This allows the policy to distinguish the first completed stage from the second, resolving the ordering ambiguity that arises when multiple keyframe embeddings are visually similar.

\textbf{Memory Injection via Cross-Attention}. Keyframe memory is injected at the visual encoding stage, before the language model processes the observation. For each camera, let $X_t \in \mathbb{R}^{N \times D}$ denote the current frame's patch tokens from SigLIP. The keyframe bank tokens are projected and concatenated across the $K$ slots to form Keys and Values: $\text{KV} = [\tilde{F}_{k_1}; \tilde{F}_{k_2}; \cdots; \tilde{F}_{k_K}] \in \mathbb{R}^{(K \cdot 16) \times D}$. Cross-attention is then applied with the current frame tokens as queries: $X'_t = \text{CrossAttn}\!\left({X_t},\; \text{KV},\; \text{mask}=\mathbf{m}\right)$, where  $\mathbf{m}$ masks padded slots. The output is fused into the current representation via a learned residual gate: $\hat{X}_t = X_t + \sigma\!\left(g\!\left([X_t;\, X'_t]\right)\right) \cdot X'_t$ , where $g(\cdot)$ is a linear projection with negative bias initialization to suppress the memory contribution at the start of training. This design ensures that early in training, when keyframe embeddings are not yet informative, the gate remains nearly closed and the base policy behavior is preserved. The updated tokens $\hat{X}_t$ are passed to the language model in place of the original image tokens, with no change to the prefix sequence length or the language model architecture.

\subsection{Keyframe-Consistent Training}
\label{sec:keyframe_training}

A key observation motivating our design is that stage-critical moments, which are stored as keyframes, are also the moments where policy errors are most consequential. An incorrect action at a stage transition propagates through all subsequent stages. We therefore up-weight the flow matching loss at keyframe-adjacent timesteps:
\begin{equation}
    \mathcal{L} = \frac{\sum_t w_t \cdot \ell_t}{\sum_t w_t}, \quad w_t = \begin{cases} \lambda & \text{if } \min_i |t - k_i| \leq \delta \\ 1 & \text{otherwise}\end{cases},
\end{equation}
where $\ell_t = \frac{1}{H}\sum_{h=1}^H \|v_t^h - u_t^h\|^2_2$ is the per-timestep flow-matching loss, $v_t^h$ and $u_t^h$ denote the predicted and target flow
velocities at horizon step $h$, respectively. $\delta$ is a window radius around each keyframe, and $\lambda > 1$ is the keyframe weight.

Crucially, the same keyframe indices that define $w_t$ also determine the contents of $\mathcal{M}_t$ at inference time. This keyframe consistency encourages the policy to be accurate near the
events that will later be stored as memory. The supervision weights and
memory contents are therefore derived from the same event indices.


\section{Experiments}
\label{sec:result}
Having devloped our KEMO, we now evaluate our method on real-world dual-arm experiments, addressing the following main research questions: \textbf{Q1}: Does event-driven keyframe selection provide advantages over alternative frame selection strategies? \textbf{Q2}: Do the individual components of our design each contribute to the final performance? \textbf{Q3}: Does our method outperform other state-of-the-art baselines on memory-dependent tasks? 

\subsection{Experimental Setup}

Our KEMO policy is initialized from the pretrained VLA backbone $\pi_{0.5}$ and fine-tuned via supervised fine-tuning (SFT) on the collected demonstration data.
\textbf{Baselines.}
We compare against two baselines: (1) \boldsymbol{$\pi_{0.5}$}~\cite{intelligence2025pi05visionlanguageactionmodelopenworld}, which is fine-tuned on the same demonstration data using the same base training recipe as KEMO, but without memory augmentation; and (2) \textbf{MemoryVLA}~\cite{shi2025memoryvla}, which serves as a memory-augmented reference baseline.

\textbf{Task Design.}
We design six challenging memory-dependent robot manipulation tasks in the real world, presented in increasing order of average trajectory duration. These tasks span from 2 to 6 scored subtasks and durations from 28 to 95 seconds which consists 830 to 2846 execution steps.
\textbf{\textit{Swap Foods}} involves exchanging food items between two plates and requires remembering their original locations.
\textbf{\textit{Find Block}} involves covering three colored blocks and later uncovering the cup containing the target block, requiring memory of the target block's location after occlusion.
\textbf{\textit{Cover Blocks}} involves covering three blocks in sequence and then uncovering them in reverse order, requiring tracking of the current task stage.
\textbf{\textit{Box Refill}} involves removing items from several boxes and later refilling each box with a matching item, requiring memory of the item--box associations.
\textbf{\textit{Make Sandwich}} involves alternating bread placement and sauce application, requiring tracking of the current assembly progress.
\textbf{\textit{Drawer Items Replacement}} involves removing items from drawers and later placing matching new items back into the corresponding drawers, requiring memory of the drawer--item associations.
Detailed task procedures, subtask definitions, and evaluation criteria are provided in the Appendix \ref{appendix:task_details}.

\begin{figure}
\centering
\includegraphics[width=1\linewidth]{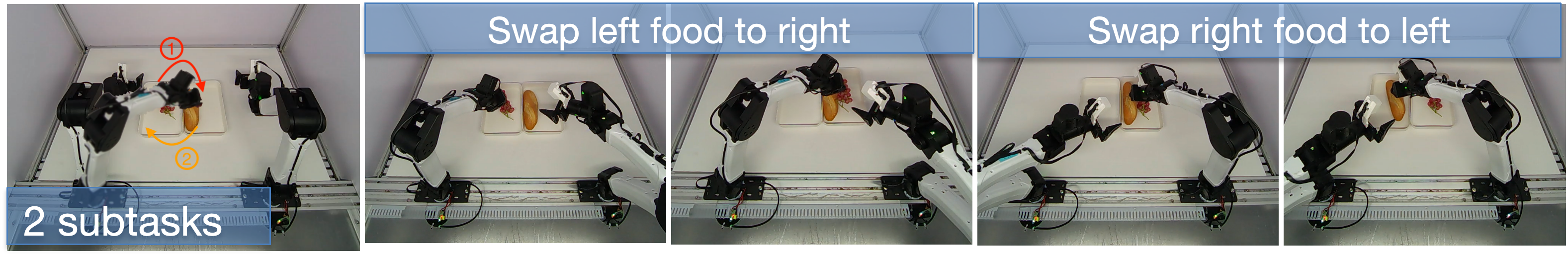}
\caption{\textbf{Swap Foods.} Trajectory length: 28 seconds (830 steps). Two food items and two plates are placed on the table. Using the proposed method, the robot exchanges the food items between the two plates: the food originally on the left plate should be moved to the right plate, and the food originally on the right plate should be moved to the left plate.}
\label{fig:task_swap_foods}
\end{figure}
\begin{figure}
\centering
\includegraphics[width=1\linewidth]{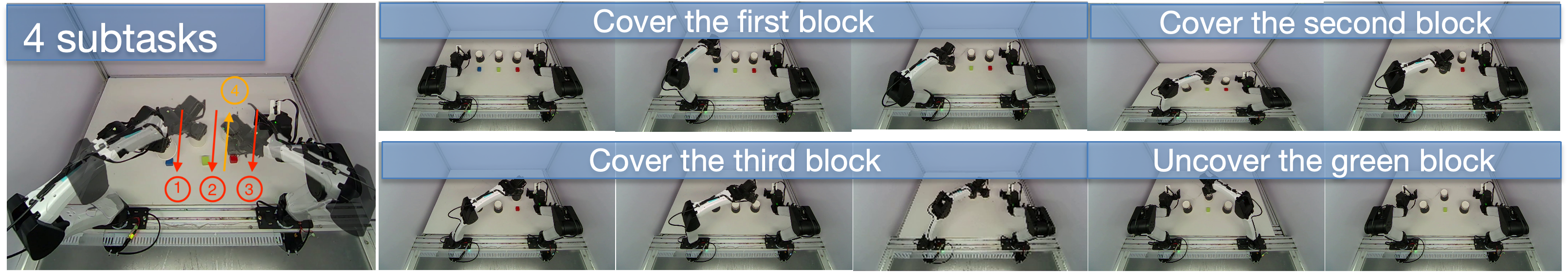}
\caption{\textbf{Find Block.} Trajectory length: 40 seconds (1186 steps).Three blocks (red, green, blue) and three cups are placed on the table. The robot sequentially covers all three blocks with cups, then identify and uncover only the cup hiding the green block, using the proposed method.}
\label{fig:task_find_block}
\end{figure}
\begin{figure}
\centering
\includegraphics[width=1\linewidth]{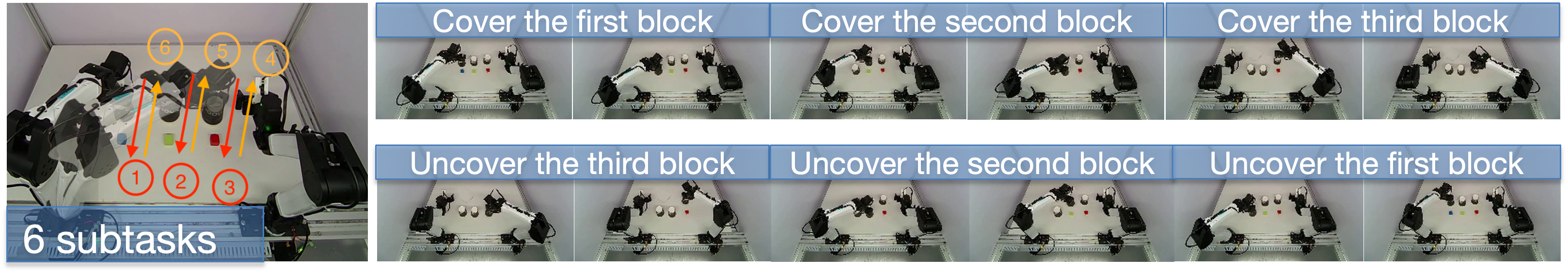}
\caption{\textbf{Cover Blocks}. Trajectory length: 50 seconds (1480 steps).The robot must sequentially cover all three blocks with cups from left to right, then uncover them in reverse order from right to left, comprising six stage transitions in total.}
\label{fig:task_cover_blocks}
\end{figure}
\begin{figure}
\centering
\includegraphics[width=1\linewidth]{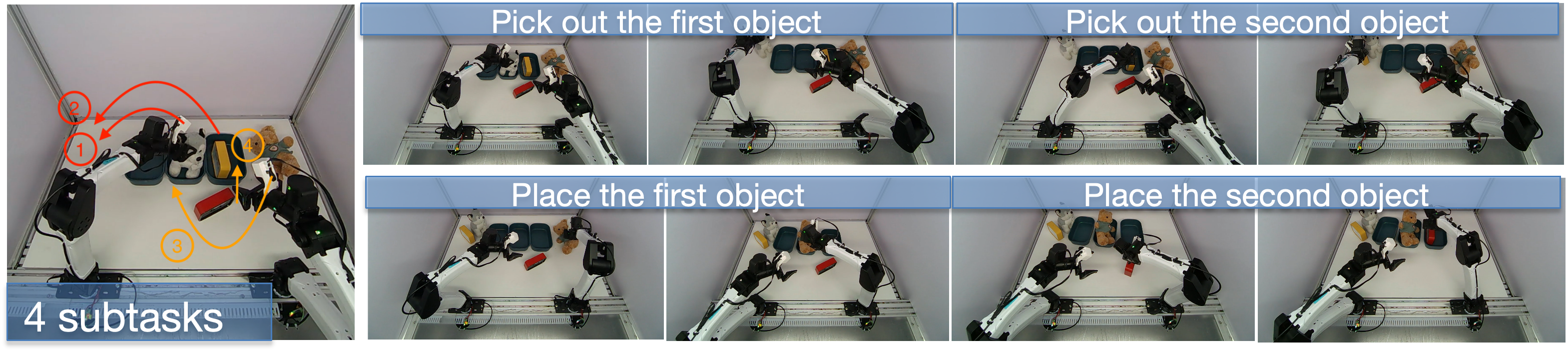}
\caption{\textbf{Box Refill.} Trajectory length: 50 seconds (1480 steps). Three boxes are placed on the table. At the beginning of each episode, two objects from two different categories are placed into two of the boxes. Using the KEMO framework, the robot first removes the original objects from the boxes and place them in the left region of the table. It then selects new objects from the right region and place each new object into the box that originally contained the same category.}
\label{fig:task_box_refill}
\end{figure}
\begin{figure}
\centering
\includegraphics[width=1\linewidth]{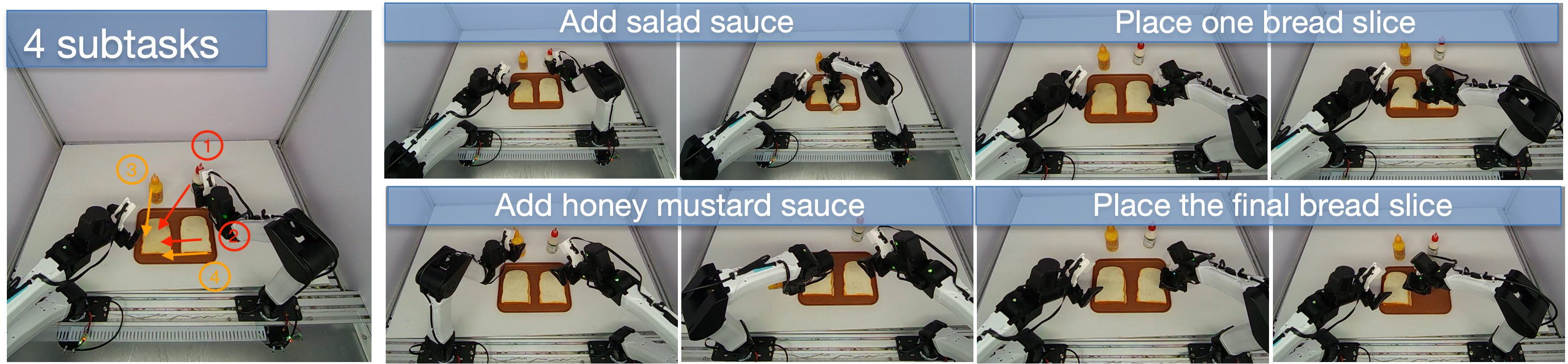}
\caption{\textbf{Make Sandwich.} Trajectory length : 54 seconds (1616 steps). Using the proposed method, the robot makes a layered sandwich by adding salad sauce, placing a bread slice, adding honey mustard sauce, and placing the final bread slice.}
\label{fig:task_make_sandwich}
\end{figure}
\begin{figure}
\centering
\includegraphics[width=1\linewidth]{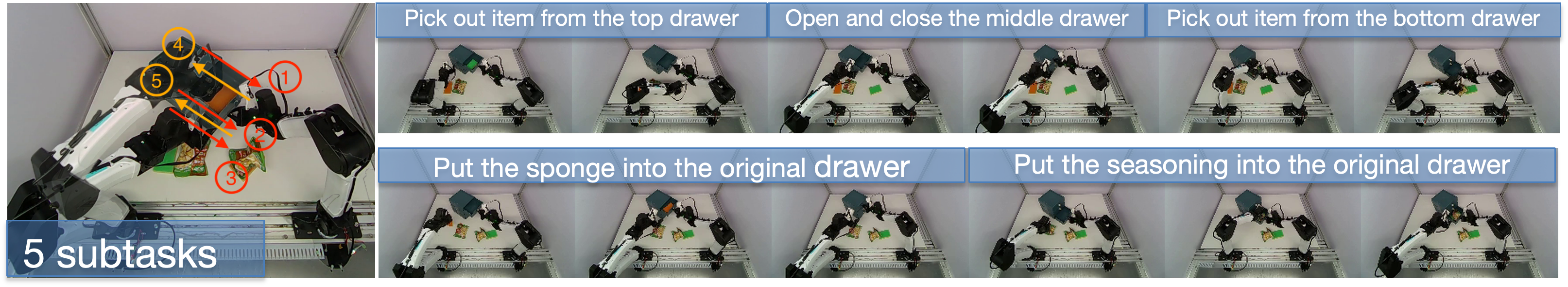}
\caption{\textbf{Drawer Items Replacement.} Trajectory length: 95 seconds (2846 steps).Two objects are placed in two of the three drawers, with their initial drawer locations randomized across episodes. Using the proposed KEMO VLA, the robot inspects the drawers from top to bottom, removes the two objects, and later places matching replacement objects into their original drawers. Depending on the initial object locations, an episode contains four or five execution subtasks. The illustrated episode contains five subtasks because the middle drawer is empty and must be inspected between the two object-containing drawers. Evaluation uses four object-relevant scored stages: two removal stages and two replacement stages.}
\label{fig:task_drawer}
\end{figure}

\textbf{Training and Evaluation Setup.}
Real-world experiments are conducted using YAM dual robot arms. Each task is trained with 100--200 demonstration data and evaluated from randomized initial states. Each method is evaluated over 12 trials using two metrics: \textbf{Task Success Rate (TSR)} and \textbf{Stage Completion Rate (SCR)}. TSR is the fraction of trials in which all task stages are completed successfully and in the correct order. For a task with $N$ stages, SCR measures the average number of consecutively completed stages from the beginning of the task, reported out of $N$. Once a stage fails, subsequent stages are not counted, even if they are later completed. Thus, TSR measures full-task completion, while SCR provides a finer-grained measure of sequential progress. Further implementation and evaluation details are provided in Appendix~\ref{appendix:implementation}.

\begin{table*}[t]
    \centering
    \caption{
    \textbf{Memory selection and component ablations on the Cover Blocks task.}
    For memory selection, retrieval and gated fusion are fixed, while
    keyframe-aligned loss weighting is disabled.
    For component ablation, event-driven keyframes are used while gated
    fusion or loss weighting is individually removed.
    TSR is reported as successful trials out of 12, and SCR as the
    average number of sequentially completed stages out of 6.
    }
    \label{tab:ablation}
    \vspace{0.6em}
    \begin{minipage}[t]{0.47\textwidth}
        \centering
        \textit{Memory Selection Strategy}\\[0.3em]
        \begin{tabular}{@{}lcc@{}}
            \toprule
            Method & TSR$\uparrow$ & SCR$\uparrow$ \\
            \midrule
            Uniform Sampling
                & 0/12
                & 1.833/6 \\
            Recent Frames
                & 1/12
                & 0.700/6 \\
            \textbf{Event Keyframes}
                & \textbf{3/12}
                & \textbf{2.500/6} \\
            \bottomrule
        \end{tabular}
    \end{minipage}
    \hfill
    \begin{minipage}[t]{0.47\textwidth}
        \centering
        \textit{Component Ablation}\\[0.3em]
        \begin{tabular}{@{}lcc@{}}
            \toprule
            Method & TSR$\uparrow$ & SCR$\uparrow$ \\
            \midrule
            w/o Gated Fusion
                & 0/12
                & 0.000/6 \\
            w/o Loss Weighting
                & 3/12
                & 2.500/6 \\
            \textbf{Ours}
                & \textbf{9/12}
                & \textbf{5.250/6} \\
            \bottomrule
        \end{tabular}
    \end{minipage}
\end{table*}

\subsection{Effect of Memory Selection Strategy}
To address Q1, we compare event-driven keyframe selection with uniform sampling and recent-frame retention on the Cover Blocks task. Retrieval and gated fusion are fixed across all variants, while keyframe-aligned loss weighting is disabled to isolate the effect of memory selection. Results are shown in Table~\ref{tab:ablation}. 
Uniform sampling selects frames at evenly spaced intervals from the observed history. However, retries during inference can lengthen the trajectory and shift the temporal positions selected by uniform sampling relative to training, resulting in inconsistent historical context. It achieves a TSR of 0/12 (0\%) and an SCR of 1.833/6 (30.6\%). The recent-frame baseline retains the $K$ most recent observations, preserving dense short-term context while discarding earlier observations. It achieves a TSR of 1/12 (8.3\%) and an SCR of 0.700/6 (11.7\%). Event-driven keyframe selection retains frames associated with action-induced task-state changes. It achieves the best performance, with a TSR of 3/12 (25.0\%) and an SCR of 2.500/6 (41.7\%). These results show that the proposed event-driven selection is more effective than uniform sampling or recent-frame retention.

\begin{table}[t]
\centering
\caption{\textbf{Performance comparison on the six real-world manipulation tasks.}
Each method is evaluated over 12 trials per task.
TSR is reported as successful trials out of 12, and SCR as the average number of sequentially completed stages over the task-specific total number of stages.}
\label{tab:Real_world}
\begin{tabular}{lcccccc}\toprule
\diagbox{Method}{Task} & \multicolumn{2}{c}{Swap Foods}&\multicolumn{2}{c}{Find Block}&\multicolumn{2}{c}{Cover Blocks}\\\midrule
 &  TSR$\uparrow$&SCR$\uparrow$&  TSR$\uparrow$&SCR$\uparrow$& TSR$\uparrow$&SCR$\uparrow$\\
$\pi_{0.5}$       &  6/12&1.500/2&  0/12&0.353/4& 0/12&1.333/6\\
MemoryVLA      &  1/12&0.750/2&  0/12&1.000/4& 0/12&0.333/6\\
\textbf{Ours}  &  \textbf{8/12}&\textbf{1.580/2}&  \textbf{2/12}&\textbf{3.167/4}& \textbf{9/12}&\textbf{5.250/6}\\ \bottomrule 
\end{tabular}

\vspace{6pt}

\begin{tabular}{lcccccc}\toprule
\diagbox{Method}{Task} & \multicolumn{2}{c}{Box Refill}&\multicolumn{2}{c}{Make Sandwich}&\multicolumn{2}{c}{\makecell{Drawer Items\\Replacement}}\\\midrule
 &  TSR$\uparrow$&SCR$\uparrow$&  TSR$\uparrow$&SCR$\uparrow$& TSR$\uparrow$&SCR$\uparrow$\\
$\pi_{0.5}$       &  4/12&2.580/4&  10/12&3.330/4& 0/12&0.000/4\\
MemoryVLA      &  0/12&0.000/4&  0/12&1.000/4& 0/12&0.000/4\\
\textbf{Ours}  &  \textbf{7/12}&\textbf{3.330/4}&  \textbf{11/12}&\textbf{3.750/4}& 0/12&\textbf{1.417/4}\\ \bottomrule 
\end{tabular}
\end{table}

\subsection{Module Contribution Analysis}

For Q2, we performed a component-wise ablation by individually disabling each module to isolate its contribution, with results in Table \ref{tab:ablation}.

\textbf{Fusion Mechanism.}
Without gating, the cross-attention output is directly added to the current visual tokens, which can cause memory features to dominate the visual representation and weaken information from the current observation. This is undesirable because action prediction still requires an accurate understanding of the robot's present state and scene configuration, while memory should serve as complementary history evidence rather than replacing the current perception. The learned gate adaptively controls the contribution of retrieved keyframe memory, allowing the policy to preserve current-observation features when they are sufficient and rely more on memory when history is needed for disambiguation.

\textbf{Loss Weighting.}
Adding keyframe-aligned loss weighting increases TSR from $3/12$ to $9/12$ and SCR from $2.50/6$ to $5.25/6$. This improvement shows that assigning greater training weight to samples near detected keyframes provides a complementary benefit to the memory mechanism and leads to more reliable sequential task execution.

Overall, the full KEMO model outperforms both ablated variants, confirming that gated memory fusion and keyframe-aligned loss weighting provide complementary benefits for long-horizon task execution.

\subsection{Comparison with Baselines}
Regarding Q3, we benchmark our method against both memory-free and memory-augmented VLA baselines across six real-world memory-dependent manipulation tasks. Table~\ref{tab:Real_world} summarizes the results. Across all tasks, $\pi_{0.5}$ achieves an aggregate TSR of  27.8\%, while MemoryVLA achieves 1.4\%. In comparison, our method achieves 51.4\%, corresponding to an absolute improvement of 23.6 percentage points over $\pi_{0.5}$. Our method also improves SCR from 42.3\% to 76.4\%, a gain of 34.1 percentage points. These results show that the proposed KEMO improves both full-task completion and sequential progress across tasks.

\section{Limitations}

	While our method demonstrates consistent improvements on memory-dependent tasks, several limitations remain. First, our method requires hyperparameter tuning, such as the sliding-window size $w$, peak-detection window, and refractory period $r$, to achieve optimal performance across different tasks. Second, the memory bank capacity $K$ is currently set manually based on the expected number of stage transitions per manipulation task. Future work will investigate learned event selection and dynamically sized memory to improve transfer across tasks.


\section{Conclusion}

In summary, we presented an KEMO framework for long-horizon memory-dependent robot manipulation. By combining kinematic and visual cues, the proposed lightweight keyframe detector identifies task-relevant events without online VLM inference or task-specific annotations while providing significant improvements over both memory-free and memory-augmented baselines. The selected keyframes provide  historical evidence for memory-augmented action prediction and are also used to emphasize event-related samples during training. Across various challenging long-horzion dual-arm manipulation tasks in the real world, our method improves aggregate TSR by 23.6\% and SCR by 34.1\% over the memory-free baseline. Ablation studies further demonstrate the benefits of event-driven keyframe selection, gated memory fusion, and keyframe-aligned loss weighting during VLA training. These results highlight event memory as an effective approach for improving sequential progress and full-task completion in long-horizon manipulation.


\clearpage
\acknowledgments{This study was supported by the InnoHK initiative of the Innovation and Technology Commission of the Hong Kong Special Administrative Region Government via the Hong Kong Centre for Logistics Robotics. The authors thank xdof.ai for supporting our data collection and real-world experiments. The authors also thank Yufeng Ji and Prof. Mengdi Xu for insightful discussions.}


\bibliography{example}  

 \clearpage
 \appendix

\section{Real-World Experimental Setup}

\begin{figure}
\centering
\includegraphics[width=0.5\linewidth]{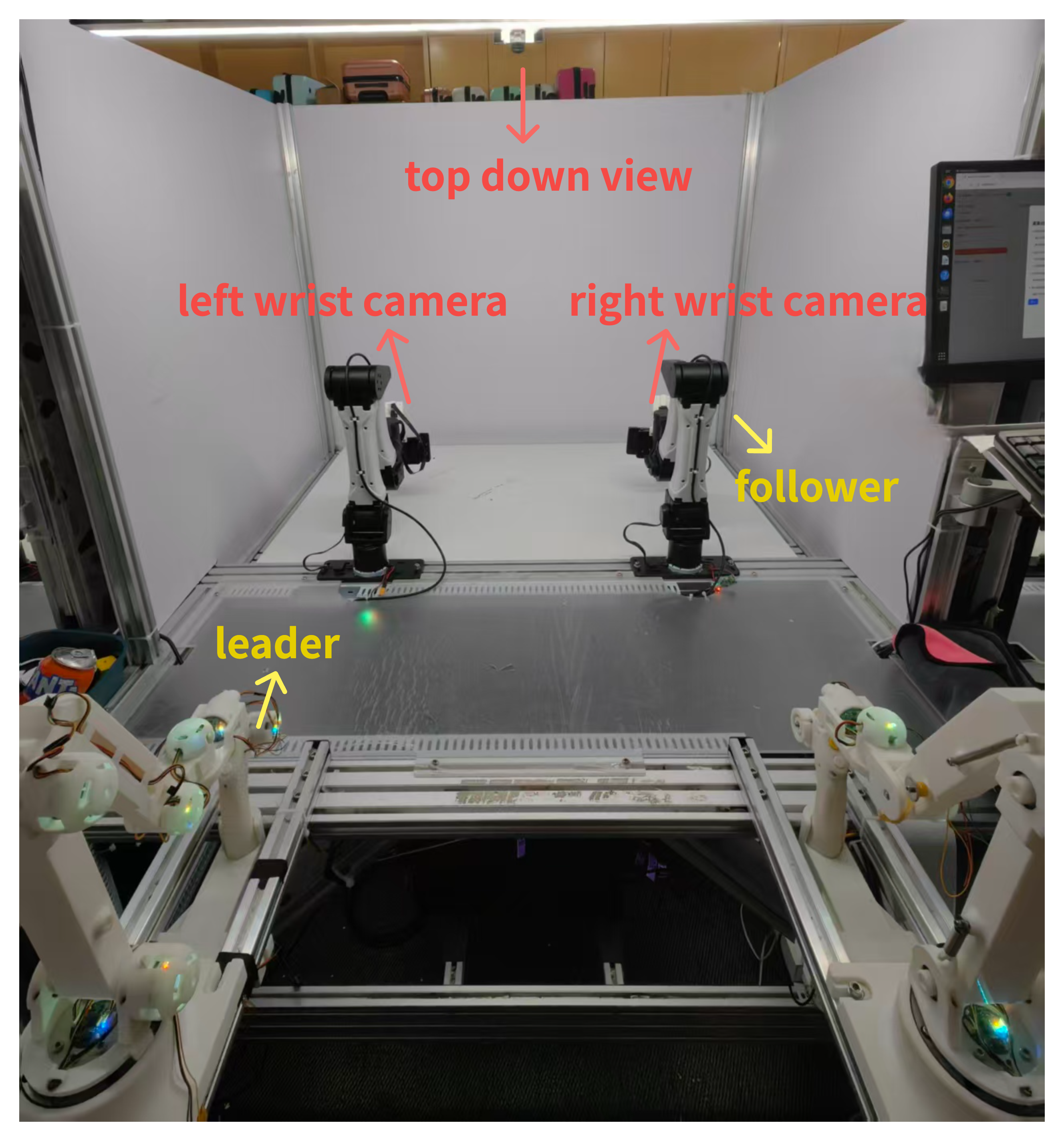}
\caption{\textbf{Real-World Experimental Setup.}Demonstrations are collected through leader--follower teleoperation, where two 6-DoF leader arms control the corresponding two 6-DoF follower arms. During policy inference and autonomous evaluation rollouts, the leader arms are not used, and only the follower arms execute the predicted actions. Visual observations are captured by three Decxin RGB cameras: one fixed top-view camera and two wrist-mounted cameras attached to the left and right follower arms.}
\label{fig:station}
\end{figure}
Figure~\ref{fig:station} shows the real-world dual-arm manipulation setup used for demonstration collection and policy evaluation. The system consists of two 6-DoF leader arms and two 6-DoF follower arms. During demonstration collection, the leader arms are used to teleoperate the corresponding follower arms. During policy evaluation, the follower arms execute the predicted actions autonomously without human intervention.

Visual observations are captured by three Decxin RGB cameras: one fixed top-view camera observing the entire workspace and two wrist-mounted cameras attached to the left and right follower arms. The top-view camera provides a global observation of the scene, while the wrist cameras provide close-range views of the manipulated objects and end-effector interactions.
 \section{Task Details}
 \label{appendix:task_details}

\subsection{Swap Foods}

\textbf{Task Description.}
As shown in Fig. \ref{fig:task_swap_foods}, two food items and two plates are placed on the table. The robot must exchange the food items between the two plates: the food originally on the left plate should be moved to the right plate, and the food originally on the right plate should be moved to the left plate.

\textbf{Language Instruction.}
Swap the two food items between the plates.

\textbf{Subtask Definition.}
\begin{itemize}
\item Subtask 1: Move the food item originally on the left plate to the right plate.
\item Subtask 2: Move the food item originally on the right plate to the left plate.
\end{itemize}

\textbf{Success Criteria.}
TSR requires both food items to be placed on the opposite plates relative to their initial positions. SCR awards one point for each food item correctly transferred to its target plate.

\textbf{Data.}
Approximately 100 demonstrations are collected via teleoperation. Each trajectory has an average duration of 28 seconds and an average length of 830 steps.


\subsection{Find Block}

\textbf{Task Description.}
As shown in Fig. \ref{fig:task_find_block}, three blocks (red, green, blue) and three cups are placed on the table. The robot must sequentially cover all three blocks with cups, then identify and uncover only the cup hiding the green block.

\textbf{Language Instruction.}
Cover the blocks from left to right using the cups, and then uncover the cup which covers the green block.

 \textbf{Subtask Definition.}
 \begin{itemize}
     \item  Subtask 1--3 (Cover phase): Place cup over each block in sequence.
     \item  Subtask 4 (Identify and uncover): Lift the cup covering the green block.

 \end{itemize}

 \textbf{Success Criteria.}
TSR requires all four subtasks to be completed correctly, including identifying the correct cup at stage 4. Uncovering the wrong cup at stage 4 counts as a failure regardless of earlier progress. SCR awards one point per correctly completed subtask.

 \textbf{Data.} 
Approximately 160 demonstrations collected via teleoperation. Each trajectory has an average duration of 40 seconds and an average length of 1186 steps.

 \subsection{Cover Blocks}
 \textbf{Task Description.}
As shown in Fig. \ref{fig:task_cover_blocks}, three blocks and three cups are placed on the table. The robot must sequentially cover all three blocks with cups from left to right, then uncover them in reverse order from right to left, comprising six stage transitions in total.

\textbf{Language Instruction.}
Cover the blocks from left to right using the lids, and then uncover them again from right to left.

 \textbf{Subtask Definition.}
 \begin{itemize}
     \item  Subtask 1--3 (Cover phase): Place cup over block 1 (left), block 2 (middle), block 3 (right), in order.
    \item  Subtask 4--6 (Uncover phase): Remove cup from block 3 (right), block 2 (middle), block 1 (left), in order.
 \end{itemize}

 \textbf{Success Criteria.}
 TSR requires all six subtasks to be completed in the correct order. SCR awards one point per correctly completed subtask in sequence; out-of-order completions do not receive credit.

 \textbf{Data.}
Approximately 200 demonstrations collected via teleoperation. Each trajectory has an average duration of 50 seconds  and an average length of 1480 steps.

\subsection{Box Refill}

\textbf{Task Description.}
As shown in Fig. \ref{fig:task_box_refill}, three boxes are placed on the table. At the beginning of each episode, two objects from two different categories are placed into two of the boxes. The robot must first remove the original objects from the boxes and place them in the left region of the table. It must then select new objects from the right region and place each new object into the box that originally contained the same category.

\textbf{Language Instruction.}
Empty the boxes, then refill each box with a new object of the same category as before.

\textbf{Subtask Definition.}
\begin{itemize}
\item Subtask 1--2 (Remove phase): Remove the two original objects from their boxes and place them in the left table region.
\item Subtask 3--4 (Refill phase): Place new objects from the right table region into the corresponding boxes according to the original box-category assignment.
\end{itemize}

\textbf{Success Criteria.}
TSR requires both boxes to be refilled with objects matching their original categories. SCR awards one point for each correctly completed remove or refill subtask in sequence. A refill subtask is counted as correct only if the new object category matches the category originally contained in that box.

\textbf{Data.}
Approximately 150 demonstrations are collected via teleoperation. Each trajectory has an average duration of 50 seconds and an average length of 1480 steps.

\subsection{Make Sandwich}

\textbf{Task Description.}
As  shown in Fig. \ref{fig:task_make_sandwich}, three bread slices and one tray are placed on the table. Initially, one bread slice is placed on the left side of the tray as the sandwich base, while two additional bread slices are stacked on the right side. The robot must make a sandwich by alternating sauce-adding and bread-stacking operations. The right arm first adds salad sauce onto the base bread and places one bread slice from the right side on top of it. The left arm then adds honey mustard sauce onto the new top bread layer, and the right arm places the final bread slice on top to complete the sandwich.

\textbf{Language Instruction.}
Make a layered sandwich by adding salad sauce, placing a bread slice, adding honey mustard sauce, and placing the final bread slice.

\textbf{Subtask Definition.}
\begin{itemize}
\item Subtask 1: The right arm adds salad sauce onto the base bread.
\item Subtask 2: The right arm places one bread slice from the right side of the tray onto the sauced base bread.
\item Subtask 3: The left arm adds honey mustard sauce onto the current top bread layer.
\item Subtask 4: The right arm places the final bread slice on top to complete the sandwich.
\end{itemize}

\textbf{Success Criteria.}
TSR requires all four subtasks to be completed in the correct order: salad sauce, bread slice, honey mustard sauce, and final bread slice. SCR awards one point for each correctly completed subtask in sequence.

\textbf{Data.}
Approximately 100 demonstrations are collected via teleoperation. Each trajectory has an average duration of 54 seconds and an average length of 1616 steps.


 \subsection{Drawer Items Replacement}

 \textbf{Task Description.}
As shown in Fig. \ref{fig:task_drawer}, one drawer unit with three drawers is  on the table. Two objects from different categories are placed in two of the three drawers, while the remaining drawer is empty. The robot must open each drawer from top to bottom, remove the item and place it on the left region of the table, then open each drawer again and place a matching new item from the right region into the correct drawer. 

 \textbf{Language Instruction.}
From top to bottom, open each drawer, pick out the item inside and remember what type of item belongs in that drawer. Then put a matching new item from the table into the same drawer.

 \textbf{Subtask Definition.}
Depending on the initial object locations, the resulting trajectory contains four or five execution subtasks. Evaluation uses four object-relevant subtasks:
 \begin{itemize}
     \item  Subtask 1--2 (Remove phase): Open each drawer top to bottom, remove item, if present, place on left table region.
     \item  Subtask 3--4 (Replace phase): Open each drawer top to bottom, place a matching new item from right table region into the corresponding drawer.
 \end{itemize}

 \textbf{Success Criteria.}
TSR requires all four object-relevant subtasks to be completed correctly: removing the two objects from their drawers and placing the two matching new objects into the corresponding drawers. SCR awards one point for each correctly completed subtask. 

 \textbf{Data.} 
 Approximately 120 demonstrations collected via teleoperation. Each trajectory has an average duration of 95 seconds and an average length of 2846 steps.

\section{Implementation Details}
\label{appendix:implementation}

\subsection{Training Configuration}
\label{Training Configuration}

\begin{table}[h]
\centering
\caption{Training hyperparameters for $\pi_{0.5}$ .  }
\begin{tabular}{lc}
\toprule
 Hyperparameter& Value \\
\midrule
 Optimizer                   & AdamW \\
   Peak learning rate          & $2.5 \times 10^{-5}$ \\
   Minimum learning rate       & $2.5 \times 10^{-6}$
\\
   Warmup steps                & 1000 \\
   Total steps                 & 30000 \\
 Batch size&32\\
 Action horizon&50\\
   Action space&  Joint-space\\
\bottomrule
\end{tabular}
\label{tab:train_config_pi}
\end{table}
\begin{table}[h]
\centering
\caption{Training hyperparameters for $\pi_{0.5}$ with event-driven keyframe memory. }
\begin{tabular}{llc}
\toprule
Module& Hyperparameter & Value \\
\midrule
 Training& Optimizer                   &AdamW \\
 & Peak learning rate          &$2.5 \times 10^{-5}$ \\
 & Minimum learning rate       &$2.5 \times 10^{-6}$\\
 & Warmup steps                &1000 \\
 & Total steps                 &30000 \\
 & Batch size&32\\
 & Action horizon&50\\
 & Action space&Joint-space\\
\midrule
Loss Weighting& Keyframe weight $\lambda$& 8.0 \\
  & Window radius $\delta
$& 3   \\
\midrule
Keyframe Memory& Spatial pooling& $4 \times4$\\
\midrule
\end{tabular}
\label{tab:train_config}
\end{table}
We follow a unified training recipe for $\pi_{0.5}$ and our method to ensure a fair comparison. Both $\pi_{0.5}$ and our method are trained for 30K steps with a global batch size of 32 on 2$\times$A100 GPUs. For visual observations, we use three camera views: a top-view camera, a left-wrist camera, and a right-wrist camera. The detailed training hyperparameters for the base policy are shown in Table~\ref{tab:train_config_pi}, and those for our memory-augmented policy are shown in Table~\ref{tab:train_config}.

\subsection{Training Details for MemoryVLA}

\begin{table}[ht]
\centering
\caption{Training hyperparameters for MemoryVLA .  }
\begin{tabular}{lc}
\toprule
 Hyperparameter& Value \\
\midrule
 Base VLM & Prismatic VLM\\
   Batch size& 128\\
   Total steps& 40000\\
   Future action window& 50\\
   Action space&  Joint-space\\
\bottomrule
\end{tabular}
\label{tab:train_config_memoryvla}
\end{table}

We trained MemoryVLA for 40K steps with a global batch size of 128 on $8 \times A100$ GPUs. Following the default MemoryVLA setting, we use a single camera as the visual input, here we use a single top-view camera. The detailed configuration is summarized in Table~\ref{tab:train_config_memoryvla}.

\section{Keyframe Detector Details}

The keyframe detector uses the same detection pipeline across tasks: it first computes an event saliency score from recent robot states, then identifies candidate peaks, and finally applies a visual deduplication filter to remove visually redundant candidates.

The detector contains three main hyperparameters: the sliding-window size $w$ for computing the motion-based event saliency score, the peak-detection window for identifying local maxima, and the refractory period $r$ for suppressing repeated detections within the same transition. The DINOv2 visual deduplication threshold is shared across tasks and set to $\epsilon = 0.05$. Task-specific values of $w$, the peak-detection window, and $r$ are summarized in Table~\ref{tab:detector_config}.

\begin{table}[h]
\centering
\caption{Task-specific keyframe detector parameters.}
\begin{tabular}{lcccccc}
\toprule
\diagbox{Parameter}{Task}& \makecell{Cover\\Blocks} & \makecell{Find\\Block} & \makecell{Drawer Items\\Replacement}& \makecell{Swap\\Foods}& \makecell{Box\\Refill}&\makecell{Make\\Sandwich}\\
\midrule
Sliding window $w$       & 30& 40& 30 & 10& 10&10\\
 Peak detection window& 20& 20&20 & 60& 60&60\\
Refractory period $r$    & 8& 8& 60 & 8& 8&8\\ \bottomrule
\end{tabular}
\label{tab:detector_config}
\end{table}

\end{document}